\pdfoutput=1
\documentclass[11pt]{article}

\usepackage[preprint]{coling}

\usepackage{times}
\usepackage{latexsym}
\usepackage{hyperref}
\usepackage{url}
\usepackage{booktabs}
\usepackage{placeins}
\usepackage{algorithm}
\usepackage{algpseudocode}
\usepackage{subcaption}
\usepackage{amsmath}
\usepackage{comment}
\usepackage{multirow}
\usepackage{makecell}
\usepackage{float}
\usepackage[toc,page]{appendix}

\usepackage[T1]{fontenc}
\usepackage[utf8]{inputenc}

\usepackage{microtype}

\usepackage{inconsolata}

\usepackage{graphicx}

\title{QE-EBM: Using Quality Estimators as Energy Loss for Machine Translation}

\author{Gahyun Yoo \\
  Department of Linguistics\\
Seoul National University\\
Seoul, Republic of Korea \\
\texttt{\{padme0421\}@snu.ac.kr} \\\And
  Jay Yoon Lee \\
 Graduate School of Data Science \\
Seoul National University \\
Seoul, Republic of Korea\\
\texttt{\{leejayyoon\}@snu.ac.kr}  \\}

\begin{document}
\maketitle
\begin{abstract}
Reinforcement learning has shown great promise in aligning language models with human preferences in a variety of text generation tasks, including machine translation. For translation tasks, rewards can easily be obtained from quality estimation (QE) models which can generate rewards for unlabeled data. Despite its usefulness, reinforcement learning cannot exploit the gradients with respect to the QE score. We propose QE-EBM, a method of employing quality estimators as trainable loss networks that can directly backpropagate to the NMT model. We examine our method on several low and high resource target languages with English as the source language. QE-EBM outperforms strong baselines such as REINFORCE and proximal policy optimization (PPO) as well as supervised fine-tuning for all target languages, especially low-resource target languages. Most notably, for English-to-Mongolian translation, our method achieves improvements of 2.5 BLEU, 7.1 COMET-KIWI, 5.3 COMET, and 6.4 XCOMET relative to the supervised baseline. 
\end{abstract}

\section{Introduction}
Reinforcement learning with human feedback (RLHF) has been used to successfully align language models with human preferences.  By assigning a high reward to responses that exhibit certain characteristics, such as lack of harmful content, we can steer the model towards generating these types of responses in various types of text generation tasks. For machine translation, reinforcement learning has been used to improve general translation quality by using quality estimation (QE) models trained on human feedback as the reward model \citep{QE_MT_RLHF, RL4LM, align-nmt-comet}. 

Quality estimation is the task of assigning a translation score to a provided source and target-side prediction pair, without referring to a gold reference. Over the last few years, quality estimation has developed rapidly, with the correlation of model ratings and human ratings of translations rising to a level at par with reference-based translation metrics \citep{wmt22-qe, wmt23-qe}. Quality estimators in the widely used COMET translation evaluation framework, such as XCOMET-QE and COMET-KIWI, use a cross-lingual encoder with a prediction head that is regressed on datasets composed of translation pairs and human ratings \citep{comet-kiwi, xcomet}.

Although reinforcement learning methods can use QE scores as rewards to produce better translations compared to vanilla fine-tuning, they fail to exploit a crucial piece of information QE scores contain: gradients. More specifically, they lose the gradients of the translation model’s parameters with respect to the QE score by treating the score as a scalar value. Energy-based training, on the other hand, can take advantage of these gradients by treating the score as a form of energy loss that can be backpropagated. 

We propose QE-EBM, a method of training a translation model by using the QE score of the model-generated translations as an energy loss. Compared to RL methods, energy-based training enables more informative distillation of knowledge from the QE model to the translation model. This method is especially useful for boosting the translation quality of low resource target languages, that do not have sufficient bilingual corpora paired with English sentences. With only the translations from the translation model, the QE model can generate scores that are a rich source of knowledge the translation models can in turn exploit through the energy loss.

There are two variants of our method: QE-STATIC and QE-DYNAMIC. In QE-DYNAMIC, the energy network's parameters are updated via contrastive learning, while in QE-STATIC, they remain fixed. The motivation for fine-tuning the energy model in QE-DYNAMIC is so that the QE model remains relevant even as the NMT model produces increasingly more natural translations similar in quality to references.

We test our method on both high resource and low-resource target languages with English as the source language. Our methods outperform supervised fine-tuning as well as REINFORCE and proximal policy optimization (PPO), showing an increase of 2.5 BLEU, 7.1 COMET-KIWI, 5.3 COMET, and 6.4 XCOMET for English-to-Mongolian translation \citep{REINFORCE, PPO}. 

\section{Related Work}
\subsection{Structured Energy Network as a Loss (SEAL)}
SEAL \citep{SEAL}  was proposed as a training framework to use energy networks as trainable loss functions for structured prediction. There are two versions of the algorithm: SEAL-STATIC and SEAL-DYNAMIC. In both versions, the task net, the neural network responsible for performing a task (i.e. prediction), is trained through a weighted sum of cross-entropy loss and energy loss. The difference lies in whether the loss net, the neural network that acts as the loss function, is updated. In SEAL-DYNAMIC, the loss net is fine-tuned with contrastive loss before each task net update so that it is better suited to predicting the energy landscape for the data samples at each step. In SEAL-STATIC, the weights of the loss net remain fixed. SEAL-DYNAMIC outperforms SPENs (Structured Prediction Energy Networks), which refers to using energy models as inference networks, as well as SEAL-STATIC in image segmentation and semantic role labeling. 

Our work can be viewed as an adaptation of SEAL to the task of translation, retaining the energy loss and iterative update algorithm while adding the following contributions: 1) we extend it to semi-supervised learning by incorporating unlabeled data, and 2) as the energy loss net, we plug in a quality estimation model which has already been fine-tuned with human annotations of translation pairs. This guides the model towards human preferences with minimal additional training and no extra data about human preference.

\subsection{Energy Based Models for Text Generation}
Existing research on using energy based models for text generation has formulated energy models as \textit{residual} energy based models \citep{ResidualEBM, joint-ebm-nlu}. Low energy samples are drawn in two steps: sampling many generations from a frozen language model, then importance sampling through the energy model. Similarly, \citet{EBR} proposes training an energy based model with a margin based loss proportional to the difference in BLEU and using it to rerank generations during inference. 

Our method has an advantage over these works in inference speed and distillation. These residual EBM methods introduces latency during inference, and do not pass the energy model's knowledge to the base language model. In contrast, we distill the knowledge of the energy model to the base NMT model during training so that we can achieve improvements even with simple greedy generation. QE-EBM is also orthogonal to these methods in that we can also use the energy network to rerank samples during inference, although we report results of greedy generation for the evaluation sets, since our method does not require sampling from the energy model during inference. 

\subsection{Using Quality Estimators in NMT Training}
\label{subsection:QE_NMT}
With the rapid development of quality estimation models, a number of recent works have incorporated quality estimators into the training of translation models. \citet{align-nmt-comet} improves translation abilities of T5-based models by applying PPO with COMET-QE as the reward model, finding that filtering training data based on the reward model is important. Using COMET-QE also, \citet{QE_MT_RLHF} improves the translation capabilities of Llama and NLLB through RAFT (Reward Ranked Fine-tuning), which involves generating multiple candidates, ranking them with the reward model, and learning from the best sample \citep{raft, llama, nllb}. 

 \citet{ReST} introduces an efficient grow-batch offline reinforcement learning algorithm consisting of alternating Grow and Improve steps. During the Grow step, samples are generated offline, and are added to the supervised batch after applying a filter based on the reward model. During the Improve step, the translation model is fine-tuned on the grown batch.

The aforementioned works of incorporating quality estimators in NMT training use a form of reinforcement learning. To the best of our knowledge, we are the first to use a quality estimation model as an energy based model and update the NMT model with gradients with respect to the score to enhance machine translation performance. 

\section{Method}

A neural machine translation system models the conditional distribution $P_{\phi}(\mathbf{y}|\mathbf{x})$ of a target sentence $\mathbf{y}$  given a source sentence $\mathbf{x}$. NMT models are trained to maximize this probability by minimizing the cross-entropy loss function, 
\begin{equation}
L_{CE}(\mathbf{x}, \mathbf{y}) = - \sum_{t=1}^{T_y} {log P_{\phi}(\mathbf{y}|\mathbf{x})}
\end{equation}

Energy-based models (EBMs) are parameterized models that output a scalar value for each input \citep{EBM}. In this paper, we investigate the use of quality estimators as EBMs, specifically COMET-KIWI \citep{comet-kiwi}. Since COMET-KIWI outputs a scalar value that represents the quality of translation for each pair of source and target sentence, the model can be represented as an energy function $E_{\theta}(\mathbf{x}, \mathbf{y}) = -s(\mathbf{x}, \mathbf{y})$ where $s$ is the COMET-KIWI score. 

To leverage the power of EBMs during training, we train the NMT model in a multi-task setup, where the loss consists of a standard cross-entropy loss and the energy term (called energy loss hereafter). For each batch consisting of $B_l$ labeled samples $\{(\mathbf{x}_{l_i}, \mathbf{y}_{l_i})\}_{i=1}^{B_l}$ and $B_u$ unlabeled samples $\{\mathbf{x}_{u_i}\}_{i=1}^{B_u}$ with $K$ translations $\{\hat{\mathbf{y}}_{u_i}^{(j)}\}_{j=1}^K$ sampled from the NMT model for each unlabeled source sentence $\mathbf{x}_{u_i}$, the loss for the NMT model can be expressed as 

\begin{equation}
\begin{split}
\label{eq:nmt_loss}
L_{NMT} = 
\alpha \cdot \frac{1}{B_l} \sum_{i=1}^{B_l}{L_{CE}(\mathbf{x}_{l_i}, \mathbf{y}_{l_i})} 
+  \\
\beta \cdot \frac{1}{B_u} \sum_{i=1}^{B_u}
\frac{1}{K}\sum_{j=1}^{K}{E_{\theta}(\mathbf{x}_{u_i},\mathbf{\hat{y}}^{(j)}_{u_i})}
\end{split}
\end{equation}

$\alpha$ is a hyper-parameter that controls the magnitude of the cross entropy gradient, and is decreased throughout training. $\beta$ is a hyper-parameter that controls the magnitude of the energy loss gradient, and is increased throughout training. More details can be found in Appendix \ref{appendix:hyperparameters}.

This hybrid loss function is one key difference between QE-EBM and previous attempts to incorporate QE into translation model fine-tuning described in Section \ref{subsection:QE_NMT}. The supervised fine-tuning and feedback training occur simultaneously, reducing the need to move back and forth between the two stages or decide on arbitrary hyperparameters such as the epoch number of each stage in each cycle.

To update the NMT model with the energy loss, we replace the regular softmax operation for the output logits from the last hidden layer of the NMT model's decoder with the straight through estimator (STE) \citep{STE}, following \citet{ENGINE}  which demonstrated success using STE to distill knowledge from an autoregressive energy network to a non-autoregressive inference network.

The exact algorithm for QE-DYNAMIC is given in Algorithm\ref{alg:qe_dynamic}. The energy network's parameters $\theta$ are updated using the labeled batch before each update of the NMT model's parameters $\phi$. The NCE loss used to train the energy network in QE-DYNAMIC is given in Equation \eqref{eq:nce_loss},

\begin{equation}
\label{eq:nce_loss}
\begin{split}
L_{E-NCE} = -\frac{1}{B_l}\sum_{i=1}^{B_l}[log \sigma(\bar{s}(\mathbf{x}_{l_i}, \mathbf{y}_{l_i})) 
+\\ \sum_{j=1}^{N}log (1-\sigma(\bar{s}(\mathbf{x}_{l_i}, \hat{\mathbf{y}}_{p_i}^{(j)})))
]
\end{split}
\end{equation}

where $\bar{s}(\mathbf{x}, \mathbf{y}) = -E_{\theta}(\mathbf{x}, \mathbf{y}) - log P_{\phi}(\mathbf{y}|\mathbf{x})$, and  $\hat{\mathbf{y}}^{(j)}_{p_i} \sim P_{\phi}(\cdot| \mathbf{x}_{l_i})$. $\sigma$ is the sigmoid function.

Through this loss, the energy model learns to prefer the gold labels over the NMT model generations.

To improve the performance of QE-STATIC and QE-DYNAMIC, we apply two data sampling techniques to select labeled and unlabeled training batches. 

\begin{itemize}
\item \textbf{Filter}: Labeled data is filtered based on COMET-KIWI score, before training begins. A labeled pair with higher COMET-KIWI score consists of a source and target sentence with higher alignment, and is likely to provide more accurate guidance for the generation model. 
\item \textbf{NN}: We select the unlabeled batch to be paired with each labeled batch by retrieving the nearest neighbor based on embedding cosine similarity between the source sentences. This is expected to reduce the mismatch in gradient direction for the model parameters that could occur as a result of training simultaneously with labeled and unlabeled data.
\end{itemize}

\begin{algorithm*}
\caption{QE-DYNAMIC Algorithm}\label{alg:cap}
\begin{algorithmic}
\Require $\{(\mathbf{x}_{l_i},\mathbf{y}_{l_i})\}_{i=1}^{B_l}, 
\{\mathbf{x}_{u_i}\}_{i=1}^{B_u}$: Training Batch
\Require $\theta$: QE model parameters
\Require $\phi$: NMT model parameters

\For{$i=1 \to B_l$}
    \Comment{Step1: sample $N$ translations for each $\mathbf{x}_{l_i}$}
    \State $ S_{l_i} \gets 
\{ \hat{\mathbf{y}}_{p_i}^{(j)} | \hat{\mathbf{y}}_{p_i}^{(j)} \sim P_{\phi}(\cdot | \mathbf{x}_{l_i}) \}_{j=1}^{N}
$
\EndFor

\State $\theta \gets \theta - \nabla_{\theta}{L_{E-NCE}}$  \Comment{Step2: update energy model}

\For{$i=1 \to B_u$}
    \State $ S_{u_i} \gets \{\hat{\mathbf{y}}_{u_i}^{(j)} | \hat{\mathbf{y}}_{u_i}^{(j)} \sim P_{\phi}(\cdot|\mathbf{x}_{u_i}) \}_{j=1}^K$ \Comment{Step3: sample K translations for $\mathbf{x}_{u_i}$}
\EndFor

\State $\phi \gets \phi - \nabla_{\phi}{L_{NMT}}$  \Comment{Step4: update NMT model}

\end{algorithmic}
\label{alg:qe_dynamic}
\end{algorithm*}

\begin{figure}[H]
        \centering
        \includegraphics[scale=0.4]{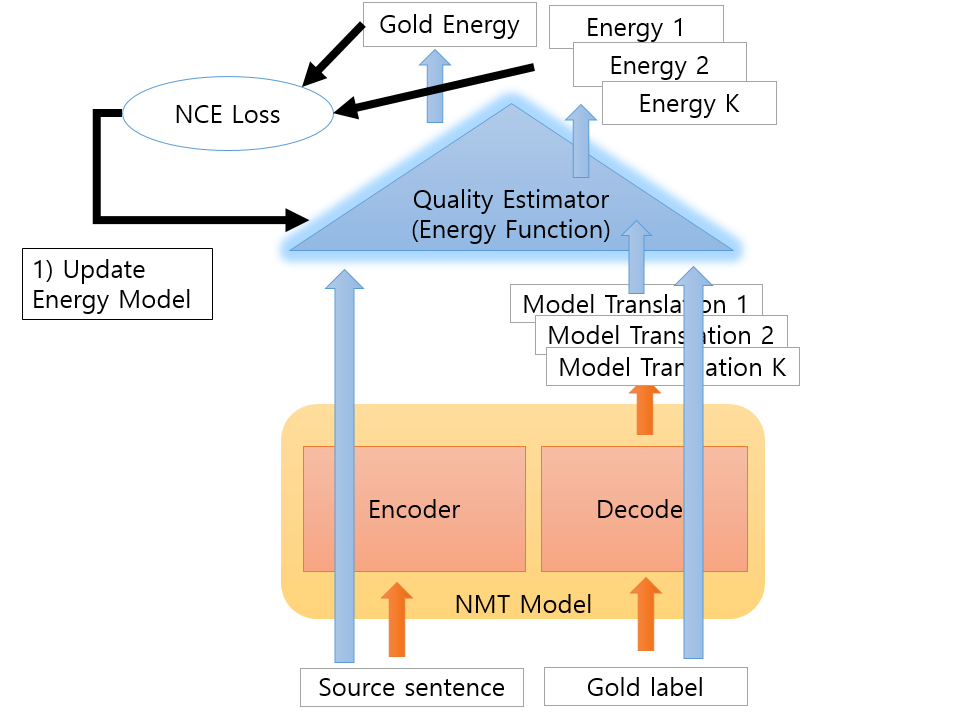}
        \caption{QE Model Update (only in QE-DYNAMIC)}
        \label{fig:qe_energy_update}
\end{figure}
\begin{figure}[H]
        \centering
        \includegraphics[scale=0.4]{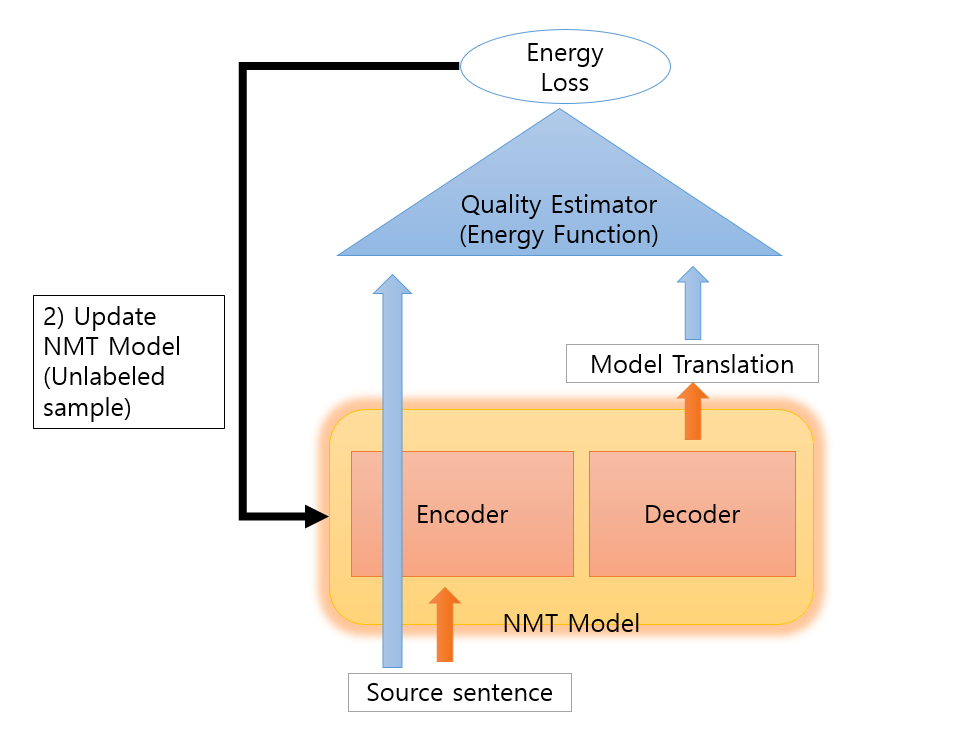}
        \caption{NMT Model Update (Unlabeled Samples)}
        \label{fig:qe_nmt_update_unlabeled}
\end{figure}

\section{Experiments}

\subsection{Models}

\paragraph{NMT Model}
For the NMT model, we use MBART, a Transformer encoder-decoder model with roughly 610M parameters \citep{mbart, Transformer}. We use the weights that are pretrained only on monolingual data and not finetuned on any parallel data, allowing us to compare regular supervised finetuning and energy-based training. 

\paragraph{QE Model}
The COMET-KIWI quality estimation model consists of a feedforward estimator on top of a cross-lingual encoder. The weights of the COMET-KIWI encoder were initialized with the InfoXLM weights and finetuned along with the feed-forward estimator on translations paired with human ratings for the WMT22 Metrics Task \citep{infoxlm}. COMET-KIWI ranked first among the reference-free metrics, and seventh among all the metrics in terms of correlation with human evaluation \citep{wmt22-qe}.

\subsection{Datasets}
We use the IWSLT2017 English-\{German, Chinese\} and ML50 English-\{Bengali, Azerbaijani, Mongolian, Marathi, Kazakh\} translation datasets \citep{iwslt17, mbart50}. In this paper, we focus on only English to X directions. We filter the training sets by total size and sentence length. For high resource language pairs, we select 50K sentence pairs randomly, and filter out sentences with more than 50 sub-word tokens in either the source sentence or reference translation. For low-resource language pairs, we only apply length-based filtering. We will refer to this processed training set as \textit{preprocessed data pool} for convenience. We construct the labeled subset by randomly selecting 1/5 of the sentence pairs (source + target sentence) in the \textit{preprocessed data pool}. For the unlabeled subset, we use all of the source sentences in the \textit{preprocessed data pool}.

\subsection{Baselines}
To the best of our knowledge, there are no previous works that use quality estimators as trainable loss functions to train translation models. Therefore, in addition to the supervised baseline,  we compare our method against REINFORCE and PPO as these reinforcement learning methods can also utilize quality estimators as a reward. Both RL baselines are trained in a multi-task setup as in QE-EBM, with the same cross-entropy loss for the labeled batch and a separate loss using the QE score for the unlabeled batch. The REINFORCE baseline uses the vanilla policy gradient. The reward is normalized through (1) scaling the reward between the running max and running min of the current epoch and (2) subtracting the running average of the current epoch. The PPO implementation is taken from TRL \citep{trl}. For a fair comparison across different approaches, we ensured an identical number of samples per source sentence in each learning algorithm. 

\subsection{Training and Evaluation}
We use an ADAM optimizer to update the energy model parameters and NMT model parameters \citep{adam}. In QE-DYNAMIC, for both the NMT model and QE model, we train adapters instead of the full parameters \citep{AdapterHub}. In QE-STATIC, we do not attach adapters to the QE model; instead we distill directly from the base model. All models were trained for 10 epochs, with early stopping based on COMET-KIWI score on the validation set. Evaluation was conducted using the metrics SacreBLEU, COMET-KIWI, COMET, and XCOMET \citep{SacreBLEU, xcomet}. COMET is the reference-based counterpart to COMET-KIWI, and XCOMET is an explainable metric that provides error spans and categories as well as sentence-level scores. More detailed experimental details can be found in Appendix \ref{appendix:exp_details}.

\subsection{Results}
We report results for QE-STATIC and QE-DYNAMIC along with the supervised baseline for low resource target languages (Bengali, Azerbaijani, Mongolian, Marathi, Kazakh) in Table \ref{tab:main_lr}. While the better-performing method varies among the two QE- variants, these methods exhibit consistent improvements over the baseline. For Mongolian, our method achieves improvements of 2.5 BLEU, 7.1 COMET-KIWI, 5.3 COMET, and 6.4 XCOMET relative to the supervised baseline. 

\begin{table}[htb]
\centering
\resizebox{0.5\textwidth}{!}{

\begin{tabular}{c c c c c} \hline
  Language Pair & \multicolumn{4}{c}{EN-BN} \\ \hline
  Data Size     & \multicolumn{4}{c}{4K} \\ \hline
  Metric        & BLEU & KIWI & COMET & XCOMET \\ \hline
  Supervised    &  5.76 & 65.46 & 65.72 & 39.64 \\
  QE-STATIC     & \textbf{6.22} & 66.66 & 65.86 & \textbf{42.33} \\ 
  QE-DYNAMIC    & 6.04 & \textbf{67.14} & \textbf{67.06} & 41.43 \\ \hline
  Language Pair & \multicolumn{4}{c}{EN-AZ} \\ \hline
  Data Size     & \multicolumn{4}{c}{5K}    \\ \hline
  Metric        & BLEU & KIWI & COMET & XCOMET \\ \hline
  Supervised    & 5.61 & 68.17 & 71.95 & 59.34 \\ 
  QE-STATIC     & 6.51 & 73.63 & \textbf{75.17} & \textbf{63.03} \\
  QE-DYNAMIC    & \textbf{6.78} & \textbf{73.65} & 75.10 & 62.96 \\ \hline
  Language Pair & \multicolumn{4}{c}{EN-MN} \\ \hline
  Data Size     & \multicolumn{4}{c}{7K} \\ \hline
  Metric        & BLEU & KIWI & COMET & XCOMET \\ \hline
  Supervised    & 4.16 & 62.97 & 70.80 & 55.26 \\ 
  QE-STATIC     & 6.62 & \textbf{70.09} & \textbf{76.10} & \textbf{61.71} \\ 
  QE-DYNAMIC    & \textbf{6.64} & 69.83 & 75.85 & 61.44 \\ \hline
  Language Pair & \multicolumn{4}{c}{EN-MR} \\ \hline
  Data Size     & \multicolumn{4}{c}{9K} \\ \hline
  Metric        & BLEU & KIWI & COMET & XCOMET \\ \hline
  Supervised    & 6.43 & 54.61 & 57.64 & 31.98 \\ 
  QE-STATIC     & 6.22 & 57.56 & 58.37 & 33.16 \\
  QE-DYNAMIC    & \textbf{6.45} & \textbf{58.44} & \textbf{58.45} & \textbf{33.26} \\ \hline
  Language Pair & \multicolumn{4}{c}{EN-KA} \\ \hline
  Data Size     & \multicolumn{4}{c}{12K} \\ \hline
  Metric        & BLEU & KIWI & COMET & XCOMET \\ \hline
  Supervised    & 9.93 & 72.47 & 73.92 & 64.64 \\ 
  QE-STATIC     & \textbf{9.93} & \textbf{73.48} & \textbf{74.48} & \textbf{65.19} \\ 
  QE-DYNAMIC    & 9.65 & 72.97 & 74.27 & 64.73 \\ \hline
\end{tabular}
}
\caption{Main results for low resource language pairs (\textbf{Bengali, Azerbaijani, Mongolian, Marathi, Kazakh}). All numbers represent the average of three experiments with different seeds. EBM methods were run with both labeled data filtering and adjacent unlabeled batch retrieval. The best result for each metric in each dataset is highlighted in bold.}
\label{tab:main_lr}
\end{table}

We report results for QE-STATIC and QE-DYNAMIC along with the supervised baseline for high resource target languages (German and Chinese) in Table \ref{tab:main_hr}. For both languages, QE-DYNAMIC outperforms QE-Static and the supervised baseline in all four metrics. The amount of improvement over supervised fine-tuning is larger for lower resource languages.

\begin{table}[htb]
\centering
\resizebox{0.5\textwidth}{!}{
\begin{tabular}{c c c c c} \hline
Language Pair & \multicolumn{4}{c}{EN-DE} \\ \hline
  Data Size   & \multicolumn{4}{c}{50K} \\ \hline
  Metric      & BLEU & KIWI & COMET & XCOMET \\ \hline
Supervised    & 24.24 & 77.23 & 79.01 & 92.07 \\
QE-STATIC     & 23.70 & 77.13 & 78.71 & 92.05 \\ 
QE-DYNAMIC    & \textbf{25.29} & \textbf{77.97} & \textbf{79.66} & \textbf{92.20} \\ \hline
Language Pair  & \multicolumn{4}{c}{EN-ZH} \\ \hline
  Data Size   & \multicolumn{4}{c}{50K} \\ \hline
  Metric      & BLEU & KIWI & COMET & XCOMET \\ \hline
Supervised    & 20.02 & 75.94 & 77.71 & 76.51 \\ 
QE-STATIC     & 20.01 & 76.29 & 77.73 & 76.14 \\
QE-DYNAMIC   & \textbf{20.10} & \textbf{76.35} & \textbf{77.88} & \textbf{76.54} \\ \hline
\end{tabular}
}
\caption{Main results for high resource language pairs (\textbf{German, Chinese}). All numbers represent the average of three experiments with different seeds. EBM methods were run with both labeled data filtering and adjacent unlabeled batch retrieval. The best result for each metric in each dataset is highlighted in bold.}
\label{tab:main_hr}
\end{table}

\section{Analysis}
In this section, we first present more comprehensive experimental results for four languages (Bengali, Marathi, German, Chinese) in Tables \ref{tab:low_detailed}, \ref{tab:high_detailed}  and \ref{tab:averaged_by_lang}. In addition to the supervised baseline and EBM methods with both data filtering and retrieval, we report the results of reinforcement learning baselines and ablation experiments. Different types of ablations were examined for each algorithm. For the REINFORCE and PPO baselines, we tried training without any additional monolingual data, using the labeled data with different shuffling as the unlabeled data (-Mono). (+Mono) refers to the original setup of using the whole \textit{preprocessed data pool} as the unlabeled data. For QE-DYNAMIC and QE-STATIC, we ran three types of ablation experiments, training without additional monolingual data (-Mono), training with only unlabeled batch retrieval (NN) or labeled batch filtering (FILTER), and training with neither NN nor FILTER.

\begin{table*}
    \centering
    \resizebox{\textwidth}{!}{
    \begin{tabular}{c c c | c c c c | c c c c} \hline 
               &       &      & \multicolumn{4}{c}{EN-BN} & \multicolumn{4}{|c}{EN-MR} \\ \hline 
               & Mono  & FILTER/NN & BLEU & KIWI & COMET & XCOMET & BLEU & KIWI & COMET & XCOMET \\ \hline 
    \multirow{2}{*}{Supervised} & - &  -  & 5.76 & 65.46 & 65.72 & 39.64 & 6.43 & 54.61 & 57.64 & 31.98 \\
                                & - & FILTER & 4.59 & 62.48 & 62.96 & 39.33 & 3.10 & 47.38 & 52.03 & 28.90 \\ \hline
    \multirow{2}{*}{REINFORCE}  & +Mono &   -  & 5.22 & 64.02 & 64.03 & 39.92 & 5.86 & 52.81 & 55.91 & 31.55 \\
               & -Mono &   -  & 5.02 & 62.53 & 63.07 & 38.22 & 5.48 & 51.53 & 55.40 & 30.84 \\ \hline 
    \multirow{2}{*}{PPO}  & +Mono &   -  & 5.53 & 64.43 & 65.09 & 40.28 & 4.77 & 49.15 & 53.98 & 29.93 \\ 
               & -Mono &   -  & 4.03 & 62.13 & 62.89 & 38.86 & 4.77 & 49.45 & 54.39 & 29.88 \\ \hline 
    \multirow{5}{*}{QE-STATIC} & \multirow{4}{*}{+Mono} &   -  & 5.56 & 64.49 & 64.94 & 39.17 & 5.47 & 52.28 & 56.68 & 30.47 \\ 
               &       & NN   & 5.79 & 65.20 & 65.60 & 40.29 & 5.92 & 54.19 & 57.01 & 31.77 \\ 
               &       & FILTER & 5.56 & 66.07 & 66.65 & 41.24 & 6.31 & 57.10 & \textbf{58.46} & 32.72 \\ 
               &       & FILTER \& NN & \textbf{6.22} & \underline{66.66} & 65.86 & \textbf{42.33} & 6.22 & \underline{57.56} & 58.37 & 33.16 \\ 
               & -Mono &   -  & 5.62 & 64.94 & 66.13 & 39.89 & 5.84 & 53.33 & 56.62 & 31.59 \\ \hline 
    \multirow{5}{*}{QE-DYNAMIC} & \multirow{4}{*}{+Mono} &   -   & 6.04 & 63.73 & 66.21 & 38.91 & 6.00 & 54.14 & 57.25 & 31.45 \\ 
               &       & NN   & 6.00 & 66.17 & \underline{66.91} & 40.78 & \textbf{6.59} & 54.48 & 57.75 & 31.92 \\ 
               &       & FILTER & 5.72 & 66.43 & 66.43 & \underline{41.59} & 6.44 & 57.46 & 57.69 & \underline{33.24} \\  
               &       & FILTER \& NN & 6.04 & \textbf{67.14} & \textbf{67.06} & 41.43 & 6.45 & \textbf{58.44} & \underline{58.45} & \textbf{33.26} \\ 
               & -Mono &   -    & \underline{6.05} & 64.59 & 65.65 & 40.22 & \underline{6.56} & 54.63 & 54.72 & 31.82 \\ \hline
    \end{tabular}
    }
    \caption{Detailed results for \textbf{Bengali, Marathi}. All numbers represent the average of three experiments with different seeds. EBM (+Mono) methods were run with all possible combinations of data-centric techniques. RL baselines were run without any data-centric techniques. The best result for each metric is highlighted in bold, and the second best is underlined.}
    \label{tab:low_detailed}
\end{table*}

\begin{table*}
    \centering
    \resizebox{\textwidth}{!}{
    \begin{tabular}{c c c | c c c c | c c c c} \hline 
                                    &      &      &  \multicolumn{4}{c}{EN-DE} & \multicolumn{4}{|c}{EN-ZH} \\ \hline 
                                    & Mono & FILTER/NN &  BLEU  &  KIWI &  COMET & XCOMET & BLEU & KIWI & COMET & XCOMET  \\ \hline 
    \multirow{2}{*}{Supervised}     &  -  &   -   &  24.24 &  77.23 &  79.01 & 92.07  & 20.02 & 75.94 & 77.71 & 76.51 \\ 
                                    &  -  &  FILTER & 23.50 & 77.11 & 78.73 & 92.01 & 19.42 & 75.17 & 77.00 & 75.81 \\ \hline
    \multirow{2}{*}{REINFORCE}      & +Mono &  -  &  24.45 &  77.21&  79.07 & 91.92  & 20.00 & 75.91 & 77.61 & 76.58 \\ 
                                    & -Mono &  -  &  24.57 & 77.36 & 79.17  & 92.16  & 20.08 & 75.82 & 77.54 & 76.08 \\ \hline 
    \multirow{2}{*}{PPO}            & +Mono &  -  &  25.41 & 77.91 & 79.73  & 92.16  & 20.05 & 75.73 & 77.60 &76.35 \\ 
                                    & -Mono &  -  & 25.36  & \underline{78.16} & \underline{79.85} & 92.27 & 20.16 & 75.79 & 77.67 & 76.22 \\ \hline 
    \multirow{5}{*}{QE-STATIC}      & \multirow{4}{*}{+Mono} &  -   & 24.31  & 77.30 & 79.10  & 92.11  & 19.90 & 75.74 & 77.51 & 76.04 \\  
                                    &      & NN  & 22.85  & 76.71  & 78.61 & 86.87 & 19.79 & 75.76 & 77.46 & 76.10 \\ 
                                    &      & FILTER & 25.16 & 77.64 & 79.32 & 92.25 & \underline{20.17} & \textbf{76.38} & \textbf{77.88} & \underline{76.60} \\
                                    &      & \makecell{FILTER \& NN} & 23.70 & 77.13 & 78.71 & 92.05 & 20.01 & 76.29 & 77.73 & 76.14 \\
                                    & -Mono &  -  & 25.47 & 77.94 & 79.79 & \underline{92.52} & 19.88 & 75.89 & 77.59 & 76.37 \\ \hline 
    \multirow{5}{*}{QE-DYNAMIC}     & \multirow{4}{*}{+Mono} &   -   & \underline{25.51} & 78.10  & 79.84 & 92.41 & 19.87 & 75.80 & 77.61 & 76.32 \\
                                    &      & NN   & \textbf{25.62} & \textbf{78.35} & \textbf{80.14} & \textbf{92.73} & 20.05 & 75.98 & 77.72 & \textbf{76.64} \\
                                    &      & FILTER & 25.48 & 77.93 & 79.65 & 92.28 & \textbf{20.27} & 76.25 & \underline{77.75} & 76.20 \\
                                    &      & \makecell{FILTER \& NN} & 25.29 & 77.97 & 79.66 & 92.20 & 20.10 & \underline{76.35} & \textbf{77.88} & 76.54 \\ 
                                    & -Mono &   -  & 25.40 & 78.08 & 79.80 & 92.34 & 19.95 & 75.71 & 77.54 & 76.54 \\ \hline
    \end{tabular}
    }
    \caption{Detailed results for \textbf{German, Chinese}. All numbers represent the average of three experiments with different seeds. EBM (+Mono) methods were run with all possible combinations of data-centric techniques. RL baselines were run without any data-centric techniques. The best result for each metric is highlighted in bold, and the second best is underlined.}
    \label{tab:high_detailed}
\end{table*}

\begin{table*}
    \centering
    \begin{tabular}{c c c | c c c c} \hline  
                                & Mono & FILTER/NN & BLEU & KIWI & COMET & XCOMET \\ \hline  
    \multirow{2}{*}{Supervised} &  -  &  -  & 14.11 & 68.31 & 70.02 & 60.05 \\ 
                                &  -  &  FILTER & 12.65 & 65.54 & 67.68 & 59.01 \\ \hline
    \multirow{2}{*}{REINFORCE} & +Mono &  -  & 13.88 & 67.49 & 69.16 & 59.99 \\ 
                                & -Mono &  -  & 13.79 & 66.81 & 68.79 & 59.33 \\ \hline 
    \multirow{2}{*}{PPO}       & +Mono &  -  & 13.94 & 66.81 & 69.10 & 59.68 \\ 
                                & -Mono &  -  & 13.58 & 66.38 & 68.70 & 59.31 \\ \hline 
    \multirow{5}{*}{QE-Static}  & \multirow{4}{*}{+Mono} &  -     & 13.81 & 67.45 & 69.56 & 59.45 \\ 
                                &                        & NN     & 13.58 & 67.96 & 69.67 & 58.76 \\  
                                &                        & FILTER & 14.30 & 69.30 & 70.57 &60.70 \\   
                                &                        & FILTER \& NN & 14.04 & 69.41 & 70.17 & \textbf{60.92} \\ 
                                & -Mono                  &   -    & 14.20 & 68.02 & 70.03 & 60.09 \\ \hline 
    \multirow{5}{*}{QE-Dynamic} & \multirow{4}{*}{+Mono} &   -    & 14.36 & 67.94 & 70.23 & 59.77 \\ 
                                &                        & NN     & \textbf{14.56} & 68.74 & \underline{70.63}& 60.52 \\ 
                                &                        & FILTER & 14.48 & \underline{69.52}& 70.38 & 60.83 \\ 
                                &                        & FILTER \& NN & 14.47 & \textbf{69.97} & \textbf{70.76} & \underline{60.86} \\ 
                                & -Mono                  &   -    & \underline{14.49}& 68.25 & 69.43 & 60.23 \\ \hline
    \end{tabular}
    \caption{Detailed results averaged for all four languages (\textbf{German, Chinese, Bengali, Marathi}). The best result for each metric is highlighted in bold, and the second best is underlined.}
    \label{tab:averaged_by_lang}
\end{table*}

\subsection{Reinforcement Learning vs Energy-based Training}
EBM methods outperform reinforcement learning methods in all metrics. We hypothesize that this superiority arises from the difference in the granularity of information the QE score holds in each algorithm. RL methods treat the score given by the quality estimation model as a simple scalar value, while EBM methods treat it as a loss that can back-propagate to update the translation model's parameters. For REINFORCE or PPO, a reward is given per sequence, meaning that it assigns the same reward for all the tokens in the same sequence. It also holds no information other than the quality of the translation pair relative to other translation pairs. Meanwhile, in the case of energy-based methods, the score holds fine-grained information about each token's contribution to the sequence-level quality estimation. The gradients with respect to the score also hold information about how each parameter of the NMT model should be updated to produce a translation that is most compatible with the source sentence. 

\subsection{QE-STATIC vs QE-DYNAMIC: Fine-tuning the QE model}
Table \ref{tab:averaged_by_lang} shows that on average, QE-DYNAMIC has higher scores compared to QE-STATIC. In QE-STATIC, It is possible for the base task model, such as our base translation model, to learn to optimize a static reward model in a way that does not improve the actual task performance.\footnote{A similar phenomenon, dubbed 'reward gaming', was reported in the reinforcement learning literature, and has been shown to occur in conditional text generation \citep{reward-gaming}.} In QE-DYNAMIC, this can be avoided by continually update the quality estimation model to make its guidance more robust.

The quality estimation model is also not explicitly trained to estimate the relative quality of several translations given the same source sentence. Instead, it is fine-tuned to regress on human ratings on a dataset of translation pairs, and thus ranks translations of different source sentences. Contrastive learning of the gold translation and model translations can improve the quality estimator's ability to make fine-grained distinctions in quality for a given source sentence. 

\subsection{Effect of Monolingual Data}
When neither FILTER nor NN is applied to QE-STATIC or QE-DYNAMIC, using monolingual data (+Mono) performs worse than not using it (-Mono). With the help of the data techniques, however, adding monolingual data surpasses not using it. This demonstrates that our methods can exploit a large amount of monolingual data in cases where parallel data is scarce to attain higher quality translation, but require tactics that help reduce the discrepancy between heterogeneous labeled and unlabeled data and ensure that the quality of parallel data is above a certain level.

\subsection{Effect of Data Sampling Techniques}
We investigate how the two data sampling techniques, labeled data filtering and adjacent unlabeled batch retrieval, contribute to the improvement of translation quality for QE-EBM.

Table \ref{tab:low_detailed} shows the effect of applying (1) nearest unlabeled batch retrieval, (2) labeled data filtering, and (3) both for each algorithm on EN-BN and EN-MR translation. For low resource target languages, applying each technique separately as well as combining them enhances translation quality for both QE-DYNAMIC and QE-STATIC. 

Table \ref{tab:high_detailed} contains comparisons for high resource languages. For both German and Chinese, applying unlabeled batch retrieval consistently improves performance for QE-DYNAMIC, but not for QE-STATIC. We hypothesize that unlabeled batch retrieval is more beneficial for QE-DYNAMIC because the energy model is fine-tuned with the labeled batch, and is likely to make the most accurate predictions on those batches. Using an unlabeled batch that is similar to the labeled batch during NMT model fine-tuning best exploits the tuned energy model.

For high resource languages, filtering mostly improves the performance of the NMT model, with the exception of QE-DYNAMIC for German.

\section{Conclusion}
We propose a new method of improving neural machine translation by using quality estimators as trainable energy loss networks. Our method performs better than  supervised fine-tuning and reinforcement learning baselines in both high and low-resource language directions, especially showing greater improvements for low-resource languages. Based on our experiments, the best performance is expected when applying QE-DYNAMIC with labeled data filtering and retrieval of nearest unlabeled batches.

\section{Limitations}
Our proposed training scheme has several limitations. First, although it reduces latency during inference compared to other methods such as energy-based reranking, the joint update requires a large amount of computation and memory during training, since we need to calculate and store gradients for both models. Second, it is difficult to use the quality estimation model as an energy model as is when there is a vocabulary mismatch between the energy model and the NMT model. This problem may be solved by adapting the quality estimator to operate in latent space, which is a potential future research direction.

\FloatBarrier

\bibliography{coling_latex}

\FloatBarrier

\appendix
\section{Hyperparameters}
\label{appendix:hyperparameters}
\begin{table}[H]
    \centering
    \begin{tabular}{|c|c|c|} \hline 
      Symbol  &  Function & Value \\ \hline 
      $\alpha$ & \makecell{weight for \\ unlabeled batch} & \makecell{$min(0, steps/1000)$ \\ $* 0.001$} \\ \hline
      $\beta$ & \makecell{weight for \\ labeled batch} & \makecell{$max(0, 1-10*\alpha)$} \\ \hline
      $N$ & \makecell{number of \\ hypotheses for \\ energy loss in \\ NMT model} & 5 \\ \hline
      $K$ & \makecell{number of \\ hypotheses for \\ contrastive learning \\ in energy model} & 5 \\ \hline
    \end{tabular}
    \caption{Hyperparameters}
    \label{tab:hyperparameters}
\end{table}

\FloatBarrier

\section{Experimental Details}
\label{appendix:exp_details}
\begin{itemize}
    \item When retrieving the nearest unlabeled batch for each labeled batch, the embeddings used to calculate cosine similarity are taken from the "all-mpnet-base-v2" checkpoint in the SBERT library \citep{sbert}.
    \item At the start of training, MBART weights are initialized to the "facebook/mbart-large-50" checkpoint from Huggingface \citep{transformers}.
    \item At the start of training, the QE model's weights are initialized to the "Unbabel/wmt22-cometkiwi-da" checkpoint from Huggingface \citep{transformers}.
    \item Adapter implementations are taken from AdapterHub \citep{AdapterHub}. 
\end{itemize}

\end{document}